\documentclass[letterpaper]{article} 
\usepackage{arxiv23}  
\usepackage{times}  
\usepackage{helvet}  
\usepackage{courier}  
\usepackage[hyphens]{url}  
\usepackage{graphicx} 
\usepackage{natbib}  
\usepackage{caption} 
\usepackage{algorithm}
\usepackage{algorithmic}

\usepackage{newfloat}
\usepackage{listings}
\DeclareCaptionStyle{ruled}{labelfont=normalfont,labelsep=colon,strut=off} 
\floatstyle{ruled}
\newfloat{listing}{tb}{lst}{}
\floatname{listing}{Listing}
\usepackage{booktabs} 
\usepackage{array}
\usepackage{relsize}
\usepackage[inline]{enumitem} 
\usepackage{varwidth}
\usepackage{multirow}
\usepackage[breaklinks,colorlinks,pagebackref=true,citecolor=blue]{hyperref}
\usepackage[labelformat=simple]{subcaption}

\usepackage{xspace}
\newcommand*{\eg}{e.g.\@\xspace}
\newcommand*{\ie}{i.e.\@\xspace}
\newcommand*{\etal}{et al.\@\xspace}
\usepackage{diagbox}

\newcolumntype{P}[1]{>{\centering\arraybackslash}p{#1}}

\title{Masked Autoencoder for Unsupervised Video Summarization}
\author {
    Minho Shim\textsuperscript{\rm 1},
    Taeoh Kim\textsuperscript{\rm 1},
    Jinhyung Kim\textsuperscript{\rm 2},
    Dongyoon Wee\textsuperscript{\rm 1}
}
\affiliations {
    \textsuperscript{\rm 1} NAVER Cloud
    \textsuperscript{\rm 2} KAIST
}

\begin{document}

\maketitle

\begin{abstract}
Summarizing a video requires a diverse understanding of the video, ranging from recognizing scenes to evaluating how much each frame is essential enough to be selected as a summary.
Self-supervised learning (SSL) is acknowledged for its robustness and flexibility to multiple downstream tasks, but the video SSL has not shown its value for dense understanding tasks like video summarization.
We claim an unsupervised autoencoder with sufficient self-supervised learning does not need any extra downstream architecture design or fine-tuning weights to be utilized as a video summarization model.
The proposed method to evaluate the importance score of each frame takes advantage of the reconstruction score of the autoencoder's decoder.
We evaluate the method in major unsupervised video summarization benchmarks to show its effectiveness under various experimental settings.
\end{abstract}

\section{Introduction}

Video summarization is the task of distilling a long video into a shorter form.
For the distillation, each frame needs to be scored based on its importance in the context of the video.
Since the definition of importance is subjective, several properties have been considered, including diversity, representativeness, and hierarchical information~\cite{zhou2018drdsn,li2018local,li2021exploring,jung2020global,zhao2017hierarchical}.
These heuristic properties are induced with several corresponding losses and a mixture of them, transforming visual features into importance scores.
We question whether any robust and fully data-driven representation exists that imposes minimal heuristic properties.

Seeking robust representation, unsupervised autoencoders for video summarization were naturally proposed~\cite{yang2015unsupervised}.
The hypothesis behind the unsupervised autoencoders is that videos in a specific domain share similar scenes, and similar parts are scored as highlights.
The autoencoders are still confined to the target domain, meaning we need a separate model for each domain of videos.
Such limitations of previous unsupervised learning methods gave rise to the need for self-supervised learning (SSL) in both image~\cite{chen2020simple,he2020momentum,grill2020bootstrap,he2022masked,bao2021beit} and video~\cite{han2020self,feichtenhofer2021large,wang2021dense,wei2022masked} domains, triggering large-scale backbones' development. 
As neither an explicit classifier nor a class label are used to train SSL models, one can expect SSL models to be more versatile for various downstream tasks.
However, the downstream tasks in the video domain is limited when excluding classification-based tasks like video action recognition, compared to natural language processing and image understanding (Sec.~\ref{sec:related_ssl}).
As no work has researched how the self-supervised knowledge of features affects unsupervised video summarization~\cite{apostolidis2021summsurvey}, we expand the topic to the video summarization task, which is a temporally dense understanding task.

In the video summarization task, some experimental settings have sufficient room to be diversified.
Existing unsupervised methods mostly use the videos in the target domain for training by spliting the target summarization dataset~\cite{jung2019csnet,apostolidis2021summsurvey}.
Alongside recent claims on the necessity of transition to new evaluation metrics~\cite{otani2019rethinking,apostolidis2020performance}, we discovered that cross-validation with split datasets in the community is not standardized and can affect each model's performance.
If a method were completely unsupervised by being agnostic to the target dataset, we could use summarization datasets as a whole for evaluation.

\begin{figure}[t]
\begin{center}
\subfloat[Separate backbone and summarization model.]{\label{fig:teaser01}
\includegraphics[width=0.8\linewidth]{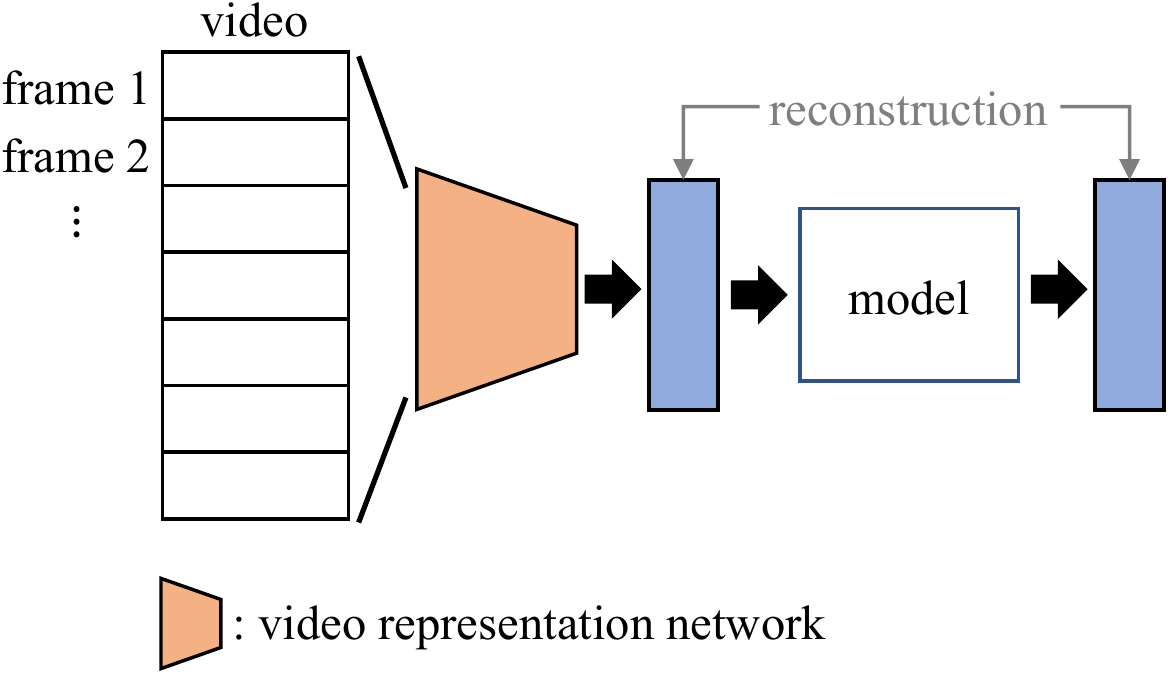}
} \\ \par\bigskip
\subfloat[Integrated backbone.]{\label{fig:teaser02}
\includegraphics[width=0.75\linewidth]{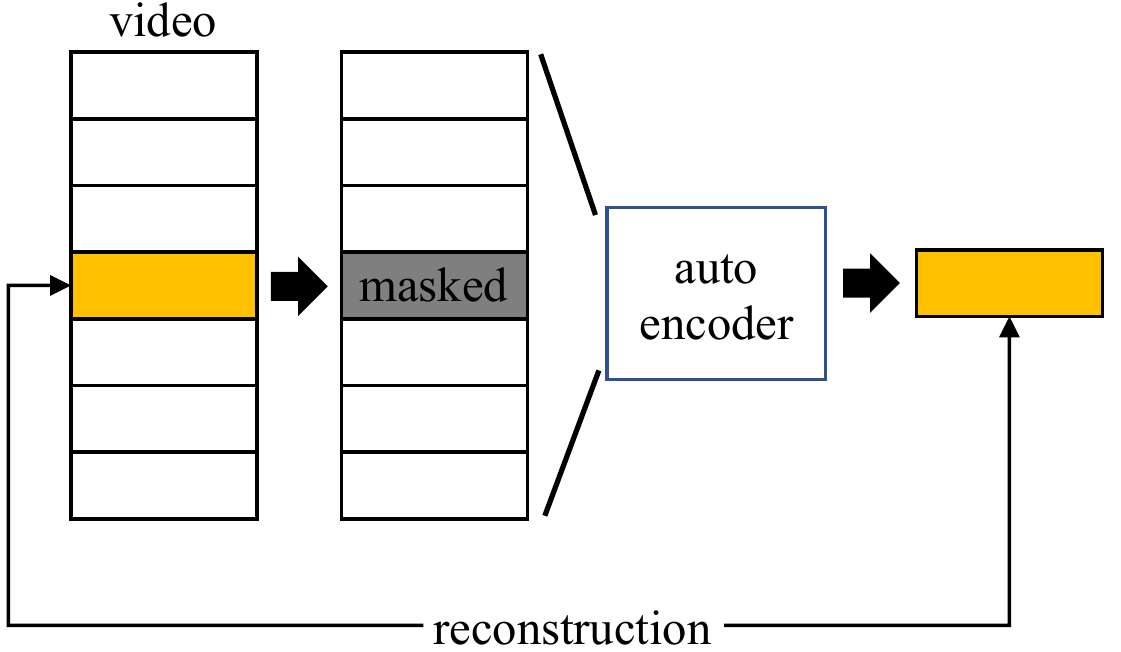}
}
\end{center}
  \caption{
  Comparison between separate backbone with summarization model, and the integrated autoencoder model.
  \protect\subref{fig:teaser01} shows a horizontal concatenation of the video representation network and a summarization model.
  The weights of the representation network is usually pretrained and fixed, and the impact of the representation is yet deeply explored.
  In \protect\subref{fig:teaser01}, \textit{reconstruction} is written in gray color to indicate it is applied when the model is an autoencoder~\cite{yang2015unsupervised}.
  Our proposed method \protect\subref{fig:teaser02} uses a single autoencoder for both pretraining and summary score prediction.
}
\label{fig:teaser}
\end{figure}

To this end, we propose to fuse a strong SSL backbone and video summarization.
Fig.~\ref{fig:teaser01} shows existing approaches that concatenate backbone models for feature extraction and separate downstream models.
In contrast, our proposed method does not require any separate summarization model, but a single autoencoder is capable of both understanding and measuring the importance score of each video frame, as shown in Fig.~\ref{fig:teaser02}.
Our autoencoder is built as a variant of masked autoencoders~\cite{he2022masked,bao2021beit}.
Recent research on masked autoencoders uses the decoder only for SSL pretraining stage, and the encoder is used alone for feature extraction or fine-tuning on video downstream tasks.
On the other hand, we propose a simple computation of decoder outputs to calculate the dissimilarity of a reconstructed frame against its original input.
Then, the dissimilarity is used as a score for evaluating each frame as a summary or not.
The process makes the proposed method completely unsupervised, \ie, labels are not used for pretraining and fine-tuning is not needed, thus the model is agnostic to summarization datasets. 

The effectiveness and simplicity of the proposed method are demonstrated in unsupervised video summarization benchmarks with minimal hyperparameters and without bells and whistles.
In addition, extensive experiments are provided to discover which aspects of pretrained video representation affect summarization quality.

In summary, our methodological, integrative, and empirical contributions are as follows:
\begin{itemize}
    \item We re-introduce autoencoder for unsupervised video summarization, with complete target domain-agnostic characteristics.
    \item To our knowledge, we provide the first bridge between self-supervised learning and video summarization. 
    \item Extensive experiments show how choices of parameters and training settings (unsupervised fine-tuning or cross-dataset validation) affect the model's effectiveness.
\end{itemize}

\section{Related Work}

\subsection{Video Summarization}

Unlike supervised learning-based deep video summarization that requires summary annotations provided by humans~\cite{zhang2016dpplstm,fajtl2018summarizing,zhao2017hierarchical,rochan2018video,zhang2018retrospective,zhao2018hsa,park2020sumgraph,saquil2021multiple,chen2022video}, unsupervised learning-based approaches have been proposed and received attention because human annotations are too subjective and expensive.
Zhou \etal~\shortcite{zhou2018drdsn} propose a reinforcement learning-based framework whose reward is based on the diversity and representativeness of the summarized video.
Another important stream of unsupervised video summarization research is to reconstruct or generate video summaries using autoencoders or generative models.
\citeauthor{yang2015unsupervised}~\shortcite{yang2015unsupervised} first proposed autoencoder-based unsupervised video summarization.
Following advances in adversarial learning~\cite{goodfellow2014generative} for various vision applications, Mahasseni~\textit{et al.}~\cite{mahasseni2017gansum} propose generators to select summary frames, and the discriminator distinguishes between the original video and the reconstructed video from summary frames.
Following works have enhanced~\cite{mahasseni2017gansum} using two-stream networks with regularization loss~\cite{jung2019csnet}, bi-directional generative models using cycle-consistency loss~\cite{yuan2019cycle}, or generative model learning from the unpaired videos~\cite{rochan2019video}.
Some works have focused on enhancing the model structure using attention mechanisms~\cite{he2019unsupervised,apostolidis2020unsupervised,jung2020global}.
Our contributions do not reside in architectural choices.
Instead, we are the first to propose a simple method to use autoencoders in a self-supervised framework for solving the video summarization task.

\begin{figure*}[t]
\begin{center}
\includegraphics[width=1.0\linewidth]{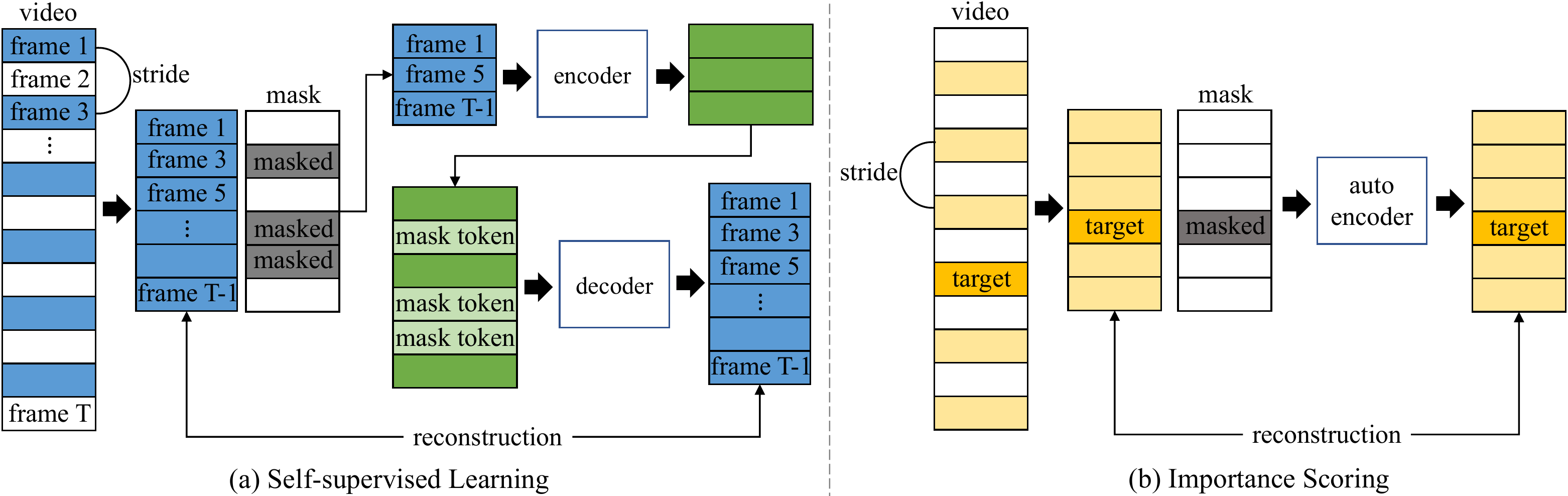}
\end{center}
  \caption{
  Design of frameworks for (a) self-supervised learning and (b) importance scoring of video summarization.
  In this figure, the temporal stride for frame sampling is set as 2 for example.
  For SSL, stride is applied based on a random starting frame.
  For summarization, stride is applied based on the target frame set as a center frame.
  }
\label{fig:architecture}
\end{figure*} 

\subsection{Self-supervised Learning} \label{sec:related_ssl}
Visual self-supervised representation learning methods~\cite{chen2020simple,he2020momentum,misra2020self,grill2020bootstrap} recently received much attention and are based on the instance discrimination~\cite{wu2018unsupervised} that learns invariance between semantically positive samples using data augmentation.
This concept is extended to the video domain~\cite{han2020self,chen2021rspnet,qian2021spatiotemporal,feichtenhofer2021large,recasens2021broaden,wang2022long}.
From the perspective of downstream tasks, image SSL has extended to object detection, segmentation~\cite{wang2021dense,roh2021spatially,henaff2021efficient}, or multi-modal tasks~\cite{huang2021seeing,kim2021vilt} combined with natural language processing which has more diverse downstream tasks~\cite{wang2018glue}.
In contrast, in the video domain, the scope of downstream tasks does not deviate significantly from action recognition~\cite{soomro2012ucf101,kuehne2011hmdb} (some studies have included fine-grained action recognition~\cite{gu2018ava,shao2020finegym}).
Recently, there has been increasing interest in masked target prediction-based SSL in the image domain~\cite{he2022masked,bao2021beit,dong2021peco,xie2022simmim} and video domain~\cite{wei2022masked}.
These methods have not only shown promising performances compared to the instance discrimination but also opened up the possibilities of new tasks beyond recognition.
Inspired by this, we propose to naturally connect the autoencoder-based method of unsupervised video summarization.

\section{Approach}

Our approach employs autoencoders for reconstructing frames and uses the reconstruction to determine the importance score of the frame.
The superficial concept is the same as the previous work using autoencoder for unsupervised video summarization~\cite{yang2015unsupervised}, but details are distinct in multiple aspects.
Early usage of the autoencoder is to find a latent space that commonly represents a given video's domain.
For example, in swimming videos, parts of videos containing the actual swimming motion are the most commonly found visual information in the videos.
Other activities are expected to be regarded as redundant parts and to be discarded.
Previous approaches~\cite{yang2015unsupervised} encode and decode video segments, searching for common segments found in each specific domain of videos.

Our model instead reconstructs a single target frame within a context of other frames.
Those other frames can be neighbors right next to the target or far from the target.
No matter the context, the autoencoder is expected to understand the relationship between the frames and reconstruct the masked target frame.

We propose a framework suitable for video self-supervised learning (SSL) and video summarization.
The autoencoder-based SSL itself is not our invention. 
We rather bridge the gap in the model architecture between the self-supervised pretraining stage and the downstream summarization task.
Existing downstream video tasks of SSL require an extra design of the head architecture built upon the SSL-trained backbone.
Also, the decoder is mostly discarded in the downstream video tasks.
Considering the cost of SSL training, reusing the decoder in the downstream task is beneficial as it has a degree of expressive power.
In order to achieve this, we introduce the method of directly using the SSL-trained encoder and decoder for summarization.

\subsection{Data Preprocessing}

Video summarization task naturally deals with long videos.
To reduce computation and memory usage, using one-dimensional features is conventional, extracted from each frame or segment of videos.
Popular choices are two-dimensional (2D) convolutional neural networks (CNN) followed by recurrent neural networks like LSTM~\cite{apostolidis2021summsurvey}.
We use 2D CNN pretrained on ImageNet to extract 1D features from all frames of videos in datasets for self-supervised learning and downstream summarization.
We do not reduce video frames per second (fps), in contrast to the recent video summarization research~\cite{zhang2016dpplstm,zhou2018drdsn,saquil2021multiple,jung2019csnet,jung2020global}. 

\subsection{Architecture} \label{sec:architecture}

Since an attention-based architecture dubbed transformer has been proposed in natural language processing~\cite{vaswani2017attention}, masked token prediction popularized the usage of the transformer architecture~\cite{devlin2018bert}.
This effort continues in computer vision by introducing vision transformers (ViT) in the image domain~\cite{dosovitskiy2020image} and its variants in the video domain~\cite{fan2021multiscale,liu2021swin}.
Both encoder and decoder in our autoencoders are built upon the transformer~\cite{devlin2018bert,dosovitskiy2020image}.
As our input is one-dimensional, the linear projection for patch embedding in ViT is not needed, while sinusoidal positional embedding is directly applied to the input sequence of frames.

\begin{table*}[t] 
\centering   
\subfloat[Fixed training stride.]{\label{tab:stride}
\setlength\extrarowheight{0.2em}
\setlength{\tabcolsep}{0.4em}
\begin{tabular}{c c | P{0.08\linewidth} P{0.08\linewidth}}
SSL & summ. & $\tau$ & $\rho$  \\
\specialrule{0.15em}{0.3em}{0.3em}
1 & 1 & 0.105 & 0.138 \\
1 & 2 & 0.105 & 0.138 \\
1 & 3 & 0.104 & 0.137 \\
1 & 4 & 0.103 & 0.135 \\
\specialrule{0.05em}{0.3em}{0.3em}
2 & 1 & 0.107 & 0.141 \\
2 & 2 & 0.107 & 0.142 \\
2 & 3 & 0.106 & 0.140 \\
2 & 4 & 0.104 & 0.137 \\
\end{tabular} } \ \ \
\subfloat[Random training stride.]{\label{tab:randomstride}
\setlength\extrarowheight{0.2em}
\setlength{\tabcolsep}{0.4em}
\begin{tabular}{c c | P{0.08\linewidth} P{0.08\linewidth}}
SSL & summ. & $\tau$ & $\rho$  \\
\specialrule{0.15em}{0.3em}{0.3em}
\textit{rand}(1,4) & 1 & 0.108 & 0.142 \\
\textit{rand}(1,4) & 2 & 0.108 & 0.142 \\
\textit{rand}(1,4) & 3 & 0.107 & 0.140 \\
\textit{rand}(1,4) & 4 & 0.105 & 0.138 \\
\specialrule{0.05em}{0.3em}{0.3em}
\underline{\textit{rand}(1,8)} & \underline{2} & 0.110 & 0.144 \\
\textit{rand}(1,8) & 4 & 0.106 & 0.140 \\
\textit{rand}(1,8) & 6 & 0.103 & 0.136 \\
\textit{rand}(1,8) & 8 & 0.101 & 0.133 \\
\end{tabular} } \\ \par\bigskip
\subfloat[Mask ratio.]{\label{tab:maskratio}
\setlength\extrarowheight{0.2em}
\setlength{\tabcolsep}{0.4em}
\begin{tabular}{c | P{0.06\linewidth} P{0.06\linewidth}}
ratio & $\tau$ & $\rho$  \\
\specialrule{0.15em}{0.3em}{0.3em}
0.10 & 0.109 & 0.144 \\
0.30 & 0.109 & 0.143 \\
\underline{0.50} & 0.110 & 0.144 \\
0.70 & 0.108 & 0.142 \\
0.90 & 0.107 & 0.140 \\
\end{tabular} } \ \ \ 
\subfloat[Input feature \& model size.]{\label{tab:feature}
\setlength\extrarowheight{0.2em}
\setlength{\tabcolsep}{0.4em}
\begin{tabular}{c c | P{0.06\linewidth} P{0.06\linewidth}}
feature & model& $\tau$ & $\rho$  \\
\specialrule{0.15em}{0.3em}{0.3em}
ResNet-50 & base & 0.084 & 0.110 \\
\underline{GoogLeNet} & \underline{base} & 0.110 & 0.144 \\
\specialrule{0.05em}{0.3em}{0.3em}
ResNet-50 & large & 0.083 & 0.110 \\
GoogLeNet & large & 0.108 & 0.143 \\
\end{tabular} } \ \ \
\subfloat[SSL epochs.]{\label{tab:epochs}
\setlength\extrarowheight{0.2em}
\setlength{\tabcolsep}{0.4em}
\begin{tabular}{c | P{0.06\linewidth} P{0.06\linewidth}}
\#ep & $\tau$ & $\rho$  \\
\specialrule{0.15em}{0.3em}{0.3em}
\underline{200} & 0.110 & 0.144 \\
400 & 0.109 & 0.143 \\
600 & 0.109 & 0.144 \\
800 & 0.109 & 0.143 \\
\end{tabular} } 
\caption[Analysis]{
Experimental results of analysis. 
Results in \protect\subref{tab:stride}-\protect\subref{tab:epochs} are trained with self-supervision on MK200 dataset and tested on TVSum dataset.
\underline{Underlined} settings are our defaults.
In \protect\subref{tab:stride}-\protect\subref{tab:randomstride}, ``SSL'' denotes values used for self-supervised training and ``summ.'' indicates values used for testing.
Higher values of Kendall's $\tau$ and Spearman's $\rho$ indicate better performance.
\label{tab:analysis} } 
\end{table*}

\subsection{Self-supervised Learning} \label{sec:ssl}

Following the recent masked target prediction-based SSL model~\cite{he2022masked}, we build an SSL model suitable for video summarization tasks with an adaptation of asymmetric encoder-decoder design.
The asymmetry means the encoder only encodes unmasked frames, while the decoder both sees encoded frames with masked frames.
Fig.~\ref{fig:architecture}(a) shows the process of SSL.
Given a sequence of frames, we randomly mask a fixed ratio of frames in the sequence.
Then, the encoder encodes the unmasked frames with positional embedding.
The output of the encoder is concatenated with masked tokens, and the order of frames is recovered.
The masked tokens consist of learnable parameters, but each mask token is identical.
The decoder reconstructs the masked regions with positional embedded inputs, and only the masked frames are used to compute the mean squared error loss.

In this SSL model, we spotlight the temporal \textit{stride} when sampling frames from the video.
We suspect the use of different strides is not studied well~\cite{wei2022masked}.
We discovered that a variable use of stride makes the summarization model more robust and flexible to different strides and improves summarization scores.
The effect of diverse strides is explored in Sec.~\ref{sec:analysis}.

\subsection{Importance Scoring for Summarization} \label{sec:summary}

Given the SSL pretrained model, the model directly predicts each video frame's importance score without additional training.
In Fig.~\ref{fig:architecture}(b), the \textit{target} is a frame targeted for scoring its importance as a summary frame.
We set the target frame to a center frame and retrieve frames around the target frame given a stride parameter.
Smaller and larger strides imply local and broad context in the perspective of the target frame.
Then, only the target frame is masked and reconstructed by the encoder-decoder of our autoencoder.
For every frame in the video, we now have a reconstructed feature and an original frame feature.
Importance is scored as a cosine dissimilarity of those two features.
The cosine dissimilarity is computed by extracting the cosine similarity of features from $1$.

\begin{table*}[t]
\centering     
\subfloat[Unsupervised fine-tuning.]{\label{tab:finetune}
\setlength\extrarowheight{0.2em}
\setlength{\tabcolsep}{0.4em}
\begin{tabular}{c c | P{0.10\columnwidth} P{0.10\columnwidth}}
\#samples & $base\_lr$ & $\tau$ & $\rho$   \\
\specialrule{0.15em}{0.3em}{0.3em}
\multicolumn{2}{c|}{\textit{w/o fine-tuning}} & 0.110 & 0.144 \\
\specialrule{0.05em}{0.3em}{0.3em}
10,000  & 4e-4 & 0.114 & 0.150  \\
50,000  & 4e-4 & 0.113 & 0.148  \\
\specialrule{0.05em}{0.3em}{0.3em}
10,000  & 4e-5 & 0.113 & 0.149  \\
50,000  & 4e-5 & 0.116 & 0.152
\end{tabular} }  \ \ \ 
\subfloat[Cross-dataset validation.]{\label{tab:crossval}
\setlength\extrarowheight{0.2em}
\setlength{\tabcolsep}{0.4em}
\begin{tabular}{P{0.24\linewidth}|c c|c c|c c}
\multirow{2}{*}{\diagbox{SSL Dataset }{Test Dataset}} &  \multicolumn{2}{c|}{TVSum} & \multicolumn{2}{c|}{SumMe} & \multicolumn{2}{c}{FineGym} \\
\cmidrule{2-7}
 & $\tau$ & $\rho$ & $\tau$ & $\rho$ & $\tau$ & $\rho$ \\
\specialrule{0.15em}{0.3em}{0.3em}
TVSum   & 0.113 & 0.149 & -0.006 & -0.007 & 0.163 & 0.200 \\
SumMe   & 0.011 & 0.014 & 0.063 & 0.077 & -0.008 & -0.009 \\
FineGym & 0.116 & 0.153 & 0.008 & 0.009 & -0.064 & -0.079
\end{tabular} } \ \ \ 
\caption[Additional analysis.]{
    Experimental results of \protect\subref{tab:finetune} unsupervised fine-tuning (Sec.~\ref{sec:finetune}) and \protect\subref{tab:crossval} cross-dataset validation (Sec.~\ref{sec:crossval}).
    TVSum dataset is used for \protect\subref{tab:finetune}.
    SSL trained models of each dataset from Table~\ref{atab:target} are used for \protect\subref{tab:crossval}.
} \label{tab:analysis2}
\end{table*}

\section{Experiments} \label{sec:experiments}

\subsection{Dataset}

For self-supervised learning, we use a video action recognition dataset Mini-Kinetics (MK200)~\cite{minikinetics}, by default.
MK200 is a balanced subset of Kinetics-400 (K400)~\cite{kinetics400}, decreasing the number of classes from 400 to 200.
Alongside K400 and MK200, we also use Kinetics-100 (K100) \cite{chen2021rspnet} with 100 classes to analyze the model's scalability.
We use MK200 by default because K400 is too large for various ablation studies and parameter searches.
As we are using Kinetics for SSL, the labels are not used.

Three benchmark datasets, TVSum~\cite{song2015tvsum}, SumMe~\cite{gygli2014summe}, and FineGym~\cite{shao2020finegym}, are used for video summarization.
For TVSum and SumMe, we use the multiple annotated importance frame-level scores as our ground truth, following previous approaches~\cite{otani2019rethinking,saquil2021multiple,apostolidis2020unsupervised}.
FineGym is originally a fine-grained action recognition dataset, but it was later proposed to be used as a summarization benchmark for longer videos~\cite{saquil2021multiple}.
Thus, there is a single summary ground truth for FineGym.
Refer to Appendix~\ref{asec:impl} for dataset properties.

\subsection{Evaluation} \label{sec:evaluation}

For several years, F1-score evaluation has been dominantly used for video summarization.
The F1-score depends on a video temporal segmentation (\eg, KTS~\cite{potapov2014category}) to generate fragments and a key-fragment selection (\eg, Knapsack)~\cite{apostolidis2021summsurvey}.
Albeit widely used, recent literature started to propose substitutes for the F1-score~\cite{otani2019rethinking,apostolidis2020performance}.
We do not mainly report F1-score for the following reasons.
First, a random prediction can get a high F1-score comparable with the SoTA unsupervised methods.
Second, even a random prediction in some cases can get a higher F1-score than human ground truth (GT) based baseline F1-score.
One can suspect F1-score may lack sufficient discrimination between random and human baseline~\cite{otani2019rethinking,apostolidis2021summsurvey}.

We use Kendall's $\tau$ and Spearman's $\rho$ to compare the ground truth scores against our model's predictions.
$\tau$ and $\rho$ both measure the association of two different ranked data, and they are proposed~\cite{otani2019rethinking} to be used for the evaluation of video summarization.
$\tau$ and $\rho$ are started to be used on the TVSum dataset~\cite{jung2020global}, and the usage is expanded over the SumMe and FineGym dataset~\cite{saquil2021multiple}.
These rank-based metrics can clearly discriminate between random and human baseline, as random predictions get zero scores and the human GT gets high scores (Table~\ref{tab:sota}).
We meticulously follow~\citeauthor{saquil2021multiple}~\shortcite{saquil2021multiple}, and use public codes and data for evaluation.

Furthermore, as our method is agnostic to the target domain dataset, training and evaluation do not require splitting data for cross-validation.
This agnostic property is beneficial because many unsupervised methods in the bibliography do not publicly open their splits, \ie, each work uses a different set of videos for cross-validation.
We simulate 5-split validation in Appendix~\ref{asec:5split}, and the results show that using a unified set or using the entire dataset like ours is crucial for a fair comparison.
Additionally, in Appendix~\ref{asec:f1score}, we share F1-scores of our models only for reference but not as our primary evaluation.

\subsection{Implementation details} \label{sec:implementation}

We extract features from video frames using ResNet-50~\cite{he2016deep} or GoogLeNet~\cite{szegedy2015going}, both pretrained on ImageNet~\cite{imagenet}, resulting in 2,048 or 1,024 dimensions of input features, respectively.
The autoencoder's encoder and decoder are two ViTs~\cite{dosovitskiy2020image} adopted for temporal features, as mentioned in Sec.~\ref{sec:architecture}.
The number of input frames is 30, and the stride for temporal frame sampling is set to a fixed value, or the stride is randomly selected for every input sample (Sec~\ref{sec:analysis}).
For example, if the temporal stride is 2, every other frame is sampled over 60 frames.
For SSL training, the start frame of a segment is randomly chosen among each video's temporal dimension with attention to the length of the video, so the end frame does not overflow the length.
Additional implementation details are in Appendix~\ref{asec:impl}.

\subsection{Analysis} \label{sec:analysis}

In this section, we explore how self-supervised training affects performances of video summarization.
Extensive experiments are conducted with different temporal strides (frame sampling rate), masking ratios, input features, and the number of training epochs.
Results are shown in Table~\ref{tab:analysis}.
Underlined settings in each table are used as our default values unless otherwise specified.
For example, the mask ratio of 0.50 in Table~\ref{tab:maskratio} is our default, and then the ratio is used for every other experiment in the other tables.
Before inspecting each property, the initial impression of values in Table~\ref{tab:analysis} implies the robustness of self-supervision.
Namely, the model is insensitive to the changes in the parameters.

\subsubsection{Temporal stride.}

In Table~\ref{tab:stride}, models are trained with a fixed value of temporal stride of 1 or 2.
Generally, as the stride used for summarization gets deviant from the stride used for training, the performance worsens.
When the model is trained with a stride of 2 and tested with strides of 2 and 4, $\tau$s are 0.107 and 0.104, respectively.
The results are intuitive as \textit{views} on each moment become contrasting if different strides are used for training and testing.

Adopting different \textit{views} on videos using multiple strides is a notable way of teaching machines to understand videos in diverse temporal contexts, \eg, SlowFast~\cite{feichtenhofer2019slowfast} of video action recognition and stride-based instance discrimination~\cite{recasens2021broaden,wang2022long}.
However, training with multiple strides is yet widely studied in the context of masked autoencoders. 
In pursuit of versatility, we propose randomly selecting a temporal stride for every training sample and letting the model reconstruct masked frames.
Even though the model is unaware of the exact stride value, the model is expected to figure out the different temporal ranges of each clip. 

In Table~\ref{tab:randomstride}, \textit{rand}(1,4) denotes that the stride value is randomly chosen from 1 to 4 for every self-supervised training sample.
The model trained with \textit{rand}(1,8) showed better performances than the model with the fixed stride.
This implies that a model with more diverse experience in temporal strides can generate better summaries.
We selected \textit{rand}(1,8) as our default training parameter, and the stride of 2 is used for default testing unless otherwise specified.
The reason for using the maximum stride of 8 is detailed in Appendix~\ref{asec:impl}.

\subsubsection{Mask ratio.}

The mask ratio determines how much proportion of frames is removed for reconstruction in a training phase.
Table~\ref{tab:maskratio} shows five different mask ratios.
Among them, the mask ratio of 0.50 gives the best performance.
The worst performing ratio is 0.90.
However, this is an interesting observation as the performance drop is only 0.003 in terms of $\tau$.
Even with most of the frames deleted, the model is capable of learning to reconstruct and generate summaries.

\subsubsection{Input feature and model size.}

We test two types of feature extraction models in Table~\ref{tab:feature}. 
ImageNet classification top-1 accuracy of the models~\cite{pytorch} are 76.13\% and 69.78\% respectively for ResNet-50 and GoogLeNet.
Even though ResNet-50's classification performance is higher, we discovered its $\tau$ and $\rho$ are significantly smaller than GoogLeNet.
With this initial impression, we try a larger model as the ResNet feature is 2048 while the GoogLeNet feature is 1024.
However, the increase in model complexity slightly drops the performance for both features.
The results suggest that features with higher classification capability are not necessarily beneficial for summarization. 
At least, the GoogLeNet feature is more suitable for our model, and we use GoogLeNet as our default input feature.

\subsubsection{Model training.} \label{sec:training}

Since SSL does not use any labels, it is hard to determine the duration of training quantitatively.
Thus, it is desired for SSL models to be robust to the number of epochs.
The robustness means the performance on downstream tasks increases to a certain point and does not decrease even if the model is SSL-trained for more epochs.
We tested four different numbers of epochs in Table~\ref{tab:epochs}.
It is worth noting that we use learning rate scheduling and warmup (Appendix~\ref{asec:impl}).
The results show our model's robustness to the number of epochs.
Our default \textit{200 epochs} experiment gives the best performance, and it is practically maintained even if the number increases up to 800.

\begin{table}[t]
\begin{center} \fontsize{9}{10}\selectfont
\begin{tabular}{@{ }l@{ }|c@{ }c|c@{ }c|c@{ }c}
\toprule
\multirow{2}{*}{Method} &  \multicolumn{2}{c|}{TVSum} & \multicolumn{2}{c|}{SumMe} & \multicolumn{2}{c}{FineGym} \\
\cmidrule{2-7}
 & $\tau$ & $\rho$ & $\tau$ & $\rho$ & $\tau$ & $\rho$ \\
\midrule
\midrule
Human & .177 & .204 & .212 & .212 & - & - \\
Random & .000 & .000 & .000 & .000 & .000 & .000 \\
\midrule
dppLSTM & .042 & .055 & -.026 & -.031 & -.027 & -.033 \\
DR-DSN & .020 & .026 & .043 & .050 & .146 & .178 \\
DR-DSN$_{2000}$ & .152 & .198 & -.016 & -.022 & NaN & NaN \\ 
GANSUM & .024 & .032 & -.010 & -.012 & - & - \\
CSNet & .025 & .034 & - & - & - & - \\
GANSUM+AAE & -.047 & -.062 & -.018 & -.023 & - & - \\
GANSUM+GL+RPE & .064 & .084 & - & - & - & -\\
CSNet+GL+RPE & .070 & .091 & - & - & - & - \\
\midrule
Ours & .110 & .144 & .046 & .057 & .055 & .067 \\
\bottomrule
\end{tabular}
\caption{
Results compared to existing unsupervised video summarization methods.
Metrics are Kendall's $\tau$ and Spearman's $\rho$.
Leading zeros are omitted.
References for each methods: dppLSTM~\cite{zhang2016dpplstm}, DR-DSN~\cite{zhou2018drdsn}, DR-DSN$_{2000}$~\cite{zhou2018drdsn,saquil2021multiple}, GANSUM~\cite{mahasseni2017gansum}, CSNet~\cite{jung2019csnet}, GANSUM+AAE~\cite{apostolidis2020unsupervised}, and GANSUM/CSNet+GL+RPE~\cite{jung2020global}.
}
\label{tab:sota}
\end{center}
\end{table}

\subsection{Unsupervised Fine-tuning} \label{sec:finetune}

The ultimate goal of SSL is to build a transdisciplinary model pretrained on a large-scale dataset.
Still, our SSL training can be directly applied to the downstream TVSum dataset for unsupervised fine-tuning.
This fine-tuning is expected to give priors about the domain of the dataset.
This kind of successive SSL training is not common, so the strategy can give interesting observations on whether the second training can boost the summarization performance.
We apply the same SSL strategy (Sec.~\ref{sec:ssl}) to the TVSum dataset.
The difference is that videos in TVSum are longer than Kinetics, so there is more room for randomly sampling a clip in the temporal dimension with the same temporal stride.

In Table~\ref{tab:finetune}, ``\#samples'' indicates the number of randomly sampled clips to be masked and reconstructed for SSL.
The experiments show that using 10,000 and 50,000 samples both effectively improves the summarization performance.
In the case of 50k samples with a base learning rate of 4e-5, $\tau$ improved by about 5.4\%.
Since 50k is much smaller than training 200 epochs on MK200, the unsupervised fine-tuning is very effective.
Also, the effortless fine-tuning implies that our SSL-trained autoencoder is a well-generalizable model for the summarization task.

\subsection{Comparison} \label{sec:comparison}

\subsubsection{Fairness.} 

Table~\ref{tab:sota} presents the results of existing unsupervised video summarization methods compared to ours.
Since the evaluation metrics have recently become popular, we bring the best values from peer-reviewed papers~\cite{jung2020global,saquil2021multiple}.
In addition, all methods except ours split the dataset for cross-validation (\eg, 5-split) and report the average of the results.
Many of them do not publicize the splits, so the settings are different for each method.
On the other hand, our method is entirely agnostic to the summarization dataset, so we evaluate our method on all videos without making any splits.
The repercussion of using 5-split validation is analyzed in Appendix~\ref{asec:5split}.

These multiple differences between methods can make the comparison look unfair.
We believe the differences have been accepted considering the subjective characteristic of the video summarization task.
As we propose our method as a new wave of video summarization, \ie as an unsupervised downstream task of SSL, Table~\ref{tab:sota} is to provide an insight into performances.
Note that the model can achieve higher scores under different settings like unsupervised fine-tuning (Sec.~\ref{sec:finetune}) and cross-dataset training (Sec~\ref{sec:crossval}).

\begin{figure}[t]
\begin{center}
\includegraphics[width=1.0\linewidth]{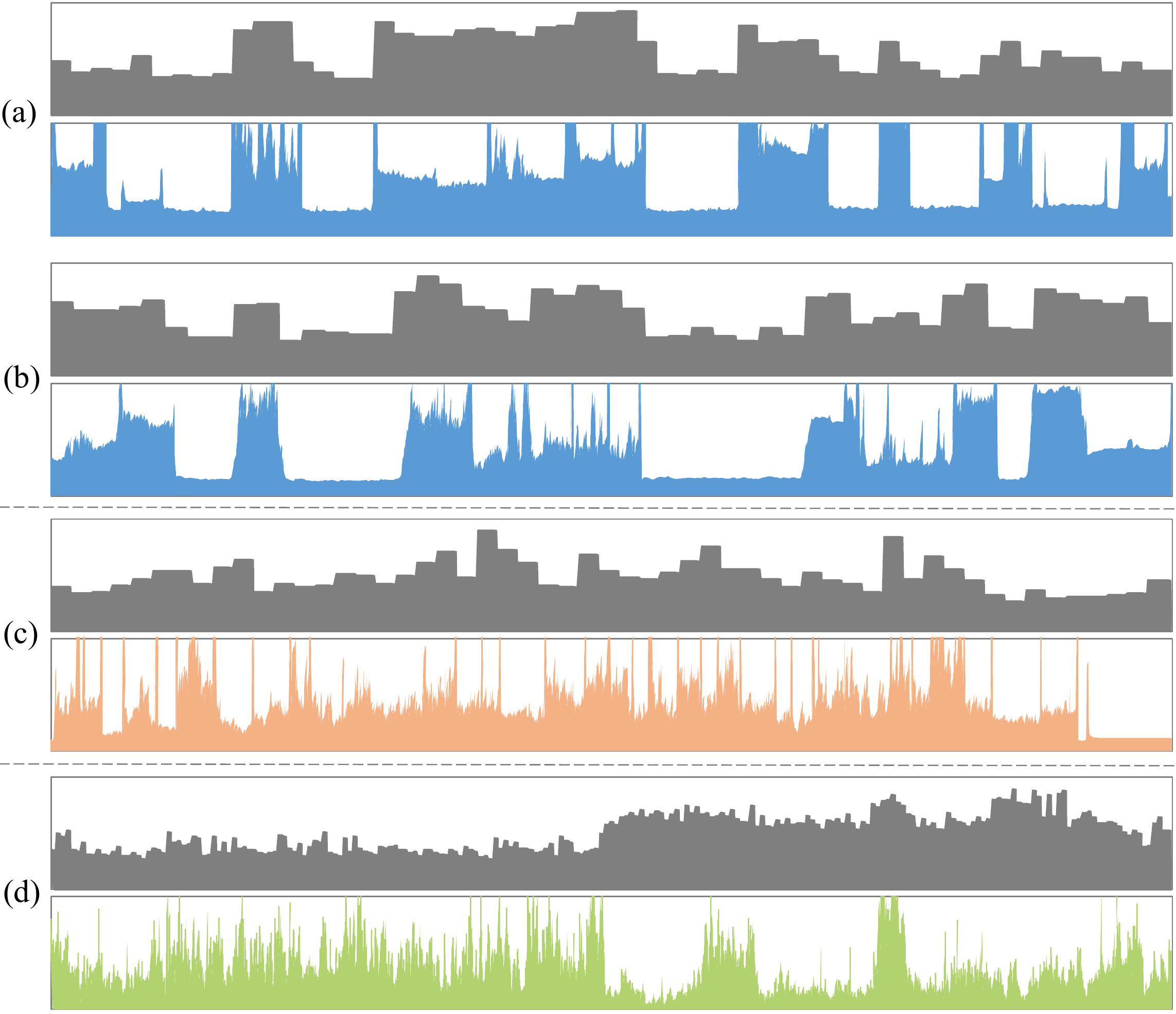}
\end{center}
 \caption{
  Visualization of ground truth and predicted importance scores.
  All charts' x-axis indicate frame number of each video, and y-axis shows scores, appropriately scaled and truncated for the sake of visualization.
  From (a) to (d), the videos (and video IDs) are video\_5 (XzYM3PfTM4w), video\_38 (EE-bNr36nyA), video\_26 (91IHQYk1IQM), and video\_33 (xmEERLqJ2kU), respectively.
  }
\label{fig:visualization}
\end{figure} 

\subsubsection{Results.}

In the TVSum dataset, the most recently proposed method in Table~\ref{tab:sota} is CSNet+GL+RPE.
Our method outperforms CSNet+GL+RPE with gains of 0.040 and 0.053 in $\tau$ and $\rho$, respectively.
In the case of the SumMe dataset, it is difficult to tell whether the evaluations are consistent with the TVSum dataset.
Three methods, dppLSTM, GANSUM, and GANSUM+AAE, show lower values of $\rho$ than the values of $\tau$.
This implies that these three methods have large deviations in their ranks' correlation against ground truth, as Spearman's $\rho$ is more sensitive to big deviations than Kendall's $\tau$.
In other words, our method and DR-DSN~\cite{zhou2018drdsn} can be seen as methods that make fewer large mistakes.
DR-DSN is notably inferior in the TVSum dataset, implying less generalization performance.
DR-DSN is originally proposed with the default of 60 epochs of training.
DR-DSN$_{2000}$ is a 2000-epoch trained version of DR-DSN later discovered by~\citeauthor{saquil2021multiple}~\shortcite{saquil2021multiple}.
DR-DSN$_{2000}$ gives high scores for TVSum, but it is reported to give unstable results for the others, showing NaN in some cases.
On the other hand, our model is robust to the number of training epochs, as shown in Table~\ref{tab:epochs}.

\subsection{Cross-dataset Validation} \label{sec:crossval}

In Sec.~\ref{sec:comparison}, we trained the model on Kinetics and tested on three datasets.
In the same manner, the model can be SSL trained on one of the three video summarization datasets and tested on each one of them.
Table~\ref{tab:crossval} shows the results.
One interesting outcome is that TVSum and FineGym are mutually complementary.
The model trained on FineGym gives the best result in TVSum testing, even better than the model trained with TVSum, and vice versa for FineGym testing.
SumMe dataset does not show any complementary characteristics with the other datasets.

\subsection{Visualization}

To help understand how our autoencoder-based summarization works, Fig.~\ref{fig:visualization} illustrates our model's predictions compared to ground truth scores.
Each of (a)-(d) is a video from the TVSum dataset, and each has two rows of charts.
The first rows with gray charts show ground truth importance scores, averaged over different annotators.
Second colored rows are charts of predicted importance scores.
(a)-(b) are strong cases, (c) is a moderate, and (d) is a failure case.

In detail, prediction score spikes are often observed to be located at the start and end of high ground truth scored regions.
As our predicted importance score is derived from reconstruction dissimilarities, high spikes are often related to shot boundaries or peculiar frames for summarization.
In the failure case of (d), it is hard to tell with our predictions which regions are worth to be highlighted, except for some distinctively high-scored parts.

\section{Conclusion and Future Work}

In this paper, we focus on introducing a new paradigm for the video summarization task.
We explore how self-supervised autoencoders can be directly used for video summarization without imposing any additional losses, fine-tuning weights, or limiting the target domain.
Our proposition is about unsupervised summarization, and supervised summarization as an SSL downstream can be the next step.
For the supervised version, it is another research area to explore if our unsupervised models with high summarization performance are aligned with the supervised models' performance, or if it is aligned with other properties or different downstream tasks like action classification.
We hope our proposed framework and analyses will establish a foothold for future work on both SSL and video summarization.

\bibliography{arxiv23}

\newpage

\setcounter{table}{0}
\renewcommand{\thetable}{A\arabic{table}}
\setcounter{figure}{0}
\renewcommand{\thefigure}{A\arabic{figure}}
\setcounter{listing}{0}
\renewcommand\thelisting{A\arabic{listing}}
\setcounter{section}{0}
\renewcommand\thesection{A\arabic{section}}

\section{Implementation Details} \label{asec:impl}

\subsubsection{Dataset properties.}

Kinetics-400 (K400) is a large-scale human-annotated video action recognition dataset. 
Each clip contains a single action class, and the duration is around 10 seconds.
As the Kinetics dataset consists of public online videos, many of them are unavailable due to expired download links.
Except unavailable videos, our K400, Mini-Kinetics-200, and Kinetics-100 contain 241,567, 77,157, and 30,570 videos for SSL training respectively.

TVSum has 50 videos, each containing 7,047 frames on average, and the score is human-annotated with an importance score by 20 different annotators.
Annotators give scores to every 2-second-long shot, and the score is applied to every frame in each shot.
The score is an integer scale from 1 to 5.
SumMe has 25 videos with 4,393 frames on average per video, and the importance score is annotated for each pre-defined shot by 15 to 18 annotators.
Every frame within a shot annotated as summary is scored 1, and the other frames are scored 0.
For FineGym, selected 50 videos have 231,603 frames on average, and the longest video spans up to 4 hours.
Segments annotated as \textit{event} timestamps are marked 1 as summary and 0 otherwise.

\subsubsection{Feature extraction.}

Each video frame is resized to $256\times256$, center-cropped to $224\times224$, and normalized with ImageNet~\cite{imagenet} statistics following standard practices~\cite{pytorch}.
Then the images are forwarded with ResNet-50~\cite{he2016deep} and GoogLeNet~\cite{szegedy2015going}.
Lanczos resampling is used for image resizing.

\subsubsection{Model.}

The encoder has 12 stacked transformer blocks, each with 12 heads for multi-head attention.
The decoder has 4 blocks, each with 6 heads.
The embedding dimension of the encoder and the decoder's attention are 768 and 384, respectively, and the embedding dimension of MLP in each block is 4 times bigger than the embedding dimension of attention (MLP ratio of 4).
Compared to the default base model, the large model (Table~\ref{tab:feature}) has bigger embedding dimensions of 1024 and 512 for the encoder and decoder, respectively.
On top of that, the encoder has 24 stacked transformer blocks with 16 heads for multi-head attention, and the decoder has 8 blocks, each with 16 heads.

\subsubsection{Self-supervised learning.}

AdamW~\cite{adamw} is used as an optimizer, the base learning rate is set to 4e-4, and linear learning rate scale rule~\cite{goyal2017accurate} ($lr=base\_lr\times batch\_size/256$) is applied alongside with 40 epochs of warmup.
Cosine learning rate schedule~\cite{loshchilov2016sgdr} is applied, and the learning rate drops to 1e-6 in the last epoch.
We do not use droppath~\cite{huang2016droppath} or gradient clipping, and all transformer blocks are initialized with Xavier uniform~\cite{glorot2010xavieruniform}.
Our default batch size is 128, and self-supervised training is done through 400 epochs.
The main reason for limiting the maximum stride value to 8 is that the Kinetics training dataset consists of 10-second videos with an average frame per second (fps) of 27.45.
Since we sample 30 frames, a stride of 8 is a safe value.
For summarization, the default stride is 2 for every experiment except SumMe, which uses the stride of 31.

\subsubsection{Unsupervised fine-tuning.}

We experimented with two different base learning rates.
One is our default, 4e-4, and the other is 4e-5.
The batch size is 128, so the practical base learning rates are 2e-4 and 2e-5 by the linear scale rule.

All three video summarization benchmarks have at most 50 videos.
This means a single clip sampled from each video cannot fill up our default batch size of 128.
For training efficiency, we sample 10 clips from each video for every training iteration.
As mentioned in Sec.~\ref{sec:finetune}, this is feasible since videos for summarization are much longer than Kinetics~\cite{kinetics400}.
The same strategy is identically applied for self-supervised learning (SSL) on summarization benchmark datasets, explained in the following Sec.~\ref{asec:ssltarget}.
Additionally, the number of warmup epochs is decreased to 5 in the case of unsupervised fine-tuning.

\subsubsection{Video action recognition.}

We describe the training procedure of downstream video action recognition explored in Sec.~\ref{sec:recognition}.
For fine-tuning and inference, the autoencoder's decoder is replaced with a single fully-connected layer. 
The layer receives a temporally average-pooled encoder's output and yields class probabilities.
All parameters of the encoder and the fully-connected layer are updated.
We train the model with a base learning rate of $0.001$, without the linear learning rate scale rule~\cite{goyal2017accurate}.
AdamW~\cite{adamw} is used as an optimizer, with betas of $(0.9, 0.999)$ and weight decay of $0.05$.
The cosine learning rate schedule~\cite{loshchilov2016sgdr} is applied with 5 warmup epochs.
The number of action recognition training epochs is $100$.
For testing, we temporally sample 4 clips from each video uniformly in the temporal dimension, and the results are averaged for action recognition.

\subsubsection{Environment.}

We use one to eight NVIDIA A100 GPUs for training and validation.
In addition, PyTorch~\cite{pytorch} library is used to build and test our architectures.

\begin{figure}[t!]
\centering     
\subfloat[TVSum.]{\label{tab:tvsumtarget}
\setlength\extrarowheight{0.2em}
\setlength{\tabcolsep}{0.4em}
\begin{tabular}{c | P{0.11\linewidth} P{0.11\linewidth} | P{0.10\linewidth} P{0.10\linewidth}}
\#samples & $\tau$ & $\rho$ & top-1 & top-5  \\
\specialrule{0.15em}{0.3em}{0.3em}
100,000 & 0.096 & 0.126 & 68.75 & 86.33 \\
200,000 & 0.109 & 0.143 & 67.72 & 85.41 \\
400,000 & 0.113 & 0.149 & 66.77 & 85.46 \\
600,000 & 0.112 & 0.147 & 67.59 & 86.20 \\
\end{tabular} } \\ \par\bigskip
\subfloat[SumMe.]{\label{tab:summetarget}
\setlength\extrarowheight{0.2em}
\setlength{\tabcolsep}{0.4em}
\begin{tabular}{c | P{0.11\linewidth} P{0.11\linewidth} | P{0.10\linewidth} P{0.10\linewidth}}
\#samples & $\tau$ & $\rho$ & top-1 & top-5  \\
\specialrule{0.15em}{0.3em}{0.3em}
50,000  & 0.056 & 0.068 & 70.39 & 88.87 \\
100,000 & 0.063 & 0.077 & 69.89 & 88.16 \\
200,000 & 0.059 & 0.073 & 70.39 & 88.87 \\
300,000 & 0.059 & 0.072 & 69.97 & 88.34 
\end{tabular} } \\ \par\bigskip
\subfloat[FineGym.]{\label{tab:finegymtarget}
\setlength\extrarowheight{0.2em}
\setlength{\tabcolsep}{0.4em}
\begin{tabular}{c | P{0.11\linewidth} P{0.11\linewidth} | P{0.10\linewidth} P{0.10\linewidth}}
\#samples & $\tau$ & $\rho$ & top-1 & top-5  \\
\specialrule{0.15em}{0.3em}{0.3em}
100,000 & -0.125 & -0.154 & 71.56 & 88.69 \\
200,000 & -0.099 & -0.122 & 69.73 & 87.95 \\
400,000 & -0.078 & -0.096 & 69.36 & 88.13 \\
600,000 & -0.064 & -0.079 & 69.10 & 87.89
\end{tabular} } 
\captionof{table}[Target domain SSL.]{
    Experimental results of conducting self-supervised learning on target domain dataset (Appendix~\ref{asec:ssltarget}). 
    Summarization evaluation metrics are Kendall's $\tau$ and Spearman's $\rho$.
    Action recognition evaluation metrics are UCF-101~\cite{soomro2012ucf101} split-1 top-1 and top-5 accuracy.
} \label{atab:target}
\end{figure}

\begin{table*}[t]
\centering     
\subfloat[Mask ratio.]{\label{tab:recognition_ratio}
\setlength\extrarowheight{0.2em}
\setlength{\tabcolsep}{0.4em}
\begin{tabular}{P{0.10\columnwidth} | P{0.10\columnwidth} P{0.10\columnwidth}}
ratio & top-1 & top-5   \\
\specialrule{0.15em}{0.3em}{0.3em}
0.10 & 71.32 & 89.21 \\
0.30 & 73.72 & 90.30 \\
0.50 & 72.09 & 89.58 \\
0.70 & 72.11 & 89.72 \\
0.90 & 72.11 & 88.29 
\end{tabular} } \ \ \ 
\subfloat[SSL training dataset.]{\label{tab:recognition_dataset}
\setlength\extrarowheight{0.2em}
\setlength{\tabcolsep}{0.4em}
\begin{tabular}{c | P{0.10\columnwidth} P{0.10\columnwidth} | P{0.10\columnwidth} P{0.10\columnwidth}}
dataset & top-1 & top-5 & $\tau$ & $\rho$  \\
\specialrule{0.15em}{0.3em}{0.3em}
Kinetics-100  & 71.40 & 88.40 & 0.110 & 0.144 \\
Mini-Kinetics-200 & 72.09 & 89.58 & 0.110 & 0.144 \\
Kinetics-400  & 73.33 & 90.72 & 0.108 & 0.142 
\end{tabular} }
\caption[Additional analysis.]{
    Experimental analysis of relationship between video action recognition and video summarization.
    TVSum dataset is used for these analyses.
} \label{tab:recognition}
\end{table*}

\section{Additional Analysis} \label{asec:add_analysis}

\subsection{Correlation with Video Action Recognition} \label{sec:recognition}

Research on video SSL treats video action recognition as a central downstream task, and the action classification performance is exploited to measure how well the knowledge provided by SSL transfers to other tasks~\cite{feichtenhofer2021large,wei2022masked}.
In order to explore if the higher action recognition scores lead to good summarization outcomes, we performed action recognition experiments on our model.
We conduct supervised fine-tuning on 1-D extracted UCF101~\cite{soomro2012ucf101} and measure split-1 accuracy.

Table~\ref{tab:recognition_ratio} shows recognition top-1 and top-5 accuracy with different mask ratios. 
In the case of summarization, Table~\ref{tab:maskratio} presented the ratio of 0.50 as the best even though the gap is small compared to other choices.
Action recognition accuracy, however, shows an apparent drop of top-1 accuracy in ratios except 0.30.
This reveals that the recognition power does not linearly imply the power of other tasks, summarization in this case.
Table~\ref{tab:recognition_dataset} presents the scalability of the model.
Action recognition accuracy increases as the size of the dataset rises from K100 to K400.
In contrast, $\tau$ and $\rho$ stay about the same regardless of the size.
We cannot decisively argue which SSL model is better with these results, but they provide initial insights into using the same backbone for diverse video understanding tasks.
Therefore, our simple but effective design of unsupervised video summarization can be another tool to find a balance between various tasks' performances.

\subsection{Self-supervision on Target Domain Dataset} \label{asec:ssltarget}

In the self-supervised part, we mainly used the Kinetics~\cite{kinetics400} dataset as it is the de facto standard for video SSL pretraining (Sec.~\ref{sec:related_ssl}).
For video summarization, three benchmark datasets are employed.
Instead, in this section, we first SSL train the model directly on video summarization datasets.
Then, the trained model is tested on the other datasets in Sec.~\ref{sec:crossval}, \eg, a model trained on TVSum~\cite{song2015tvsum} is tested on FineGym~\cite{shao2020finegym}.

Unsupervised fine-tuning on the target domain dataset is explored in Sec.~\ref{sec:finetune}.
As the method is fully unsupervised, the training scheme can be directly applied to the video summarization dataset without pretraining on Kinetics.
Since the models are trained from scratch, we increase the number of training samples, and the number of warmup epochs is set to 40.
We also conduct the same UCF-101 action recognition benchmark (Sec.~\ref{sec:recognition}).

The results are presented in Table~\ref{atab:target}.
In the case of unsupervised training on the TVSum dataset in Table~\ref{tab:tvsumtarget}, the summarization results reach their peak with 400k training samples and decrease with more samples.
However, action recognition top-1 accuracy is the worst with 400k samples.
Even the model with the best recognition accuracy is far inferior to models trained with Kinetics.
These results suggest the importance of the size and diversity of the dataset in the SSL training phase.
When trained on the SumMe dataset, the mean $\tau$ values are lower, and the mean $\rho$ values are bigger than the model trained on the Kinetics dataset.

Among three different datasets, FineGym~\cite{shao2020finegym} showed the best action recognition top-1 accuracy when trained with 100k samples (Table~\ref{tab:finegymtarget}).
This is reasonable as FineGym is at first proposed as a fine-grained action recognition dataset.
In contrast, the recognition accuracies slowly decrease as training samples increase.
Summarization scores are in the opposite situation, as $\tau$ and $\rho$ are both showing negative scores.

\section{Supplementary Results} \label{asec:supplresults}

\subsection{F1-Score} \label{asec:f1score}

Table~\ref{atab:f1} shows the key-fragment-based F1-score evaluation with different pretraining datasets.
In Sec.~\ref{sec:evaluation}, we mentioned why we do not use F1-score evaluation.
Alongside recent trends of changing evaluation metrics (some papers do not report F1-score at all)~\cite{jung2020global,saquil2021multiple}, we report these values only for references.

\begin{table}[t]
\begin{center}
\begin{tabular}{p{0.35\linewidth}|P{0.15\linewidth}|P{0.15\linewidth}}
\toprule 
Pretraining Dataset & TVSum & SumMe \\ 
\midrule 
\midrule
MK200 & 55.1 & 41.8 \\
\midrule
TVSum & 56.2 & 42.7 \\
SumMe & 54.8 & 40.6 \\
FineGym & 55.2 & 40.7 \\
\bottomrule
\end{tabular}
\end{center}
\caption{
F1-scores with different pretraining datasets.
Beside MK200, refer to Appendix~\ref{asec:add_analysis} for pretraining settings. 
FineGym does not have public data for F1-score computation.}
\label{atab:f1}
\end{table}

\subsection{Random 5-split} \label{asec:5split}

As our default method is self-supervised and agnostic to target domain datasets, we use the entire benchmark datasets for evaluation.
On the other hand, previous unsupervised methods split dataset for both training and evaluation.
The concern is that only a few previous works publicized their splits, and the others do not specify which set is used for evaluation.
It means each work is likely to use a different set of videos for training and evaluation.
In Table~\ref{atab:randomsplit}, we compare how random splits can affect the results.
We make our own random splits (\textit{Ours} in the table) and bring publicly available sets from AAE~\cite{apostolidis2020unsupervised}.
We would like to ensure that we only ran the code for random sampling five times to obtain the five splits of TVSum, SumMe, and FineGym, respectively specified in Listing~\ref{alst:tvsumsplit},~\ref{alst:summesplit}, and~\ref{alst:finegymsplit}.

Using our default model, we simulate the 5-split evaluation.
The results in Table~\ref{atab:randomsplit} show that randomness severely affects the evaluation.
Compared to \textit{Ours} in Table~\ref{tab:sota}, using the entire set for evaluation, results are very variable.
Even if the community shares a specific set for evaluation, it may still be an unfair comparison as each method can have different set that the method works well.
Also, F1-scores fail to show any performance difference in TVSum.
This analysis further strengthens the value of our proposed unsupervised and target dataset-agnostic method.

\begin{table}[t]
\begin{center} \fontsize{9}{10}\selectfont
\begin{tabular}{p{0.065\linewidth}|P{0.055\linewidth}|P{0.055\linewidth}|P{0.055\linewidth}|P{0.08\linewidth}|P{0.055\linewidth}|P{0.055\linewidth}|P{0.055\linewidth}|P{0.055\linewidth}}
\toprule
\multirow{2}{*}{Split} & 
\multicolumn{3}{c|}{TVSum} & \multicolumn{3}{c|}{SumMe} & \multicolumn{2}{c}{FineGym} \\
\cmidrule{2-9}
\cmidrule{2-9}
& $\tau$ & $\rho$ & f1 & $\tau$ & $\rho$ & f1 & $\tau$ & $\rho$ \\
\midrule
Ours & .129 & .170 & 55.8 & .028 & .035 & 40.0 & .068 & .084 \\
AAE  & .096 & .126 & 55.9 & .062 & .076 & 46.1 & - & - \\
\bottomrule
\end{tabular}
\end{center}
\caption{
Results with random 5-split test sets.
Leading zeros are omitted to save space.
AAE~\cite{apostolidis2020unsupervised} has no splits for FineGym.}
\label{atab:randomsplit}
\end{table}

\begin{listing}[tb]
\caption{Our 5-split for TVSum}
\label{alst:tvsumsplit}
\begin{lstlisting}[]
Split 1:
  video_35
  video_23
  video_15
  video_40
  video_12
  video_29
  video_21
  video_28
  video_22
  video_6
Split 2:
  video_8
  video_5
  video_22
  video_32
  video_49
  video_41
  video_21
  video_38
  video_17
  video_15
Split 3:
  video_7
  video_41
  video_5
  video_18
  video_34
  video_23
  video_24
  video_37
  video_49
  video_10
Split 4:
  video_14
  video_38
  video_16
  video_3
  video_39
  video_2
  video_46
  video_8
  video_49
  video_36
Split 5:
  video_30
  video_16
  video_32
  video_39
  video_9
  video_24
  video_42
  video_21
  video_20
  video_44
\end{lstlisting}
\end{listing}

\begin{listing}[tb]
\caption{Our 5-split for SumMe}
\label{alst:summesplit}
\begin{lstlisting}[]
Split 1:
  video_25
  video_5
  video_23
  video_20
  video_11
Split 2:
  video_12
  video_10
  video_22
  video_7
  video_15
Split 3:
  video_6
  video_13
  video_16
  video_18
  video_15
Split 4:
  video_7
  video_21
  video_2
  video_14
  video_16
Split 5:
  video_12
  video_17
  video_16
  video_22
  video_18
\end{lstlisting}
\end{listing}

\begin{listing}[tb]
\caption{Our 5-split for FineGym}
\label{alst:finegymsplit}
\begin{lstlisting}[]
Split 1:
  video_12
  video_29
  video_30
  video_26
  video_20
  video_8
  video_27
  video_40
  video_36
  video_6
Split 2:
  video_22
  video_27
  video_23
  video_5
  video_12
  video_13
  video_21
  video_44
  video_9
  video_11
Split 3:
  video_30
  video_10
  video_32
  video_27
  video_16
  video_25
  video_21
  video_15
  video_22
  video_43
Split 4:
  video_5
  video_36
  video_45
  video_44
  video_7
  video_4
  video_14
  video_42
  video_22
  video_48
Split 5:
  video_3
  video_32
  video_35
  video_18
  video_11
  video_50
  video_7
  video_41
  video_34
  video_8
\end{lstlisting}
\end{listing}

\end{document}